\documentclass[sigconf]{acmart}
\usepackage{subcaption}
\usepackage{multirow}
\AtBeginDocument{%
  \providecommand\BibTeX{{%
    \normalfont B\kern-0.5em{\scshape i\kern-0.25em b}\kern-0.8em\TeX}}}

\setcopyright{acmcopyright}
\copyrightyear{2018}
\acmYear{2018}
\acmDOI{XXXXXXX.XXXXXXX}

\begin{document}

\title{Exploring Transformers for \\On-Line Handwritten Signature Verification}

\settopmatter{printacmref=false} % Removes citation information below abstract
\renewcommand\footnotetextcopyrightpermission[1]{} % removes footnote with conference information in first column
\pagestyle{plain}

\author{Pietro Melzi,$^1$ Ruben Tolosana,$^1$ Ruben Vera-Rodriguez,$^1$ Paula Delgado-Santos,$^{1,2}$ \\Giuseppe Stragapede,$^1$ Julian Fierrez,$^1$ Javier Ortega-Garcia$^1$}
\affiliation{%
  \institution{$^1$ Universidad Autonoma de Madrid, Spain}
  \country{}
}
\affiliation{%
  \institution{$^2$ University of Kent, UK}
  \country{}
}

\renewcommand{\shortauthors}{Melzi, et al.}

\begin{abstract}
    The application of mobile biometrics as a user-friendly authentication method has increased in the last years. Recent studies have proposed novel behavioral biometric recognition systems based on Transformers, which currently outperform the state of the art in several application scenarios. On-line handwritten signature verification aims to verify the identity of subjects, based on their biometric signatures acquired using electronic devices such as tablets or smartphones. This paper investigates the suitability of architectures based on recent Transformers for on-line signature verification. In particular, four different configurations are studied, two of them rely on the Vanilla Transformer encoder, and the two others have been successfully applied to the tasks of gait and activity recognition. We evaluate the four proposed configurations according to the experimental protocol proposed in the SVC-onGoing competition. %With the simplest configuration based on the Vanilla Transformer encoder, the Equal Error Rate is reduced from 4.08\% to 3.83\%. 
    The results obtained in our experiments are promising, and promote the use of Transformers for on-line signature verification.
\end{abstract}


%\keywords{on-line signature verification, biometric, transformers, deep learning}


%\received{}
%\received[revised]{}
%\received[accepted]{}

\maketitle

\section{Introduction}
On-line handwritten signature verification is a biometric modality that aims to verify the authenticity of subjects based on their personal signatures. Handwritten signatures have been traditionally used for biometric personal recognition, as they contain unique behavioral patterns that can serve as reliable identifiers \cite{diaz2019perspective, faundez2020handwriting}. % Initially, the concept of signature verification relied heavily on manual inspection, where human experts analyzed signatures for authenticity. With the advent of digital systems, automated methods started to emerge. 
Early approaches focused on extracting dynamic or local features \cite{martinez2014mobile, tolosana2015preprocessing}, such as pen pressure, stroke sequences, speed, and acceleration, and leveraging machine learning and pattern recognition techniques, such as Dynamic Time Warping (DTW) \cite{tolosana2015preprocessing, tolosana2020exploiting} and Hidden Markov Models (HMM) \cite{fierrez2007hmm, tolosana2019reducing}. With the integration of Deep Learning (DL) \cite{tolosana2021deepsign, jiang2022dsdtw, tolosana2021deepwritesyn, tolosana2020biotouchpass2}, on-line signature verification systems have achieved remarkable performance, exhibiting robustness against various forms of forgeries \cite{tolosana2019presentation} and improving user experience \cite{tolosana2022svc}. %By leveraging advanced algorithms and machine learning techniques, on-line signature verification systems can accurately verify the identity of users, ensuring secure access to digital services while enhancing user convenience. 
This technology has wide-ranging applications in areas such as e-commerce, digital banking, and document verification, contributing to the prevention of identity fraud and improving the overall security of on-line transactions \cite{faundez2020handwriting, tolosana2017benchmarking}.

Despite the success of DL approches based on Convolutional Neural Networks (CNNs) and Recurrent Neural Networks (RNNs) \cite{tolosana2022svc}, recent studies have explored the application of Transformer architectures for other behavioral biometric traits such as gait and keystroke, outperforming the state of the art \cite{delgado2022exploring, stragapede2022typeformer, delgado2023mgaitformer}. Among the multiple advantages of Transformers, we highlight the ability to capture long-term dependencies and interactions, which is especially attractive for time series modeling \cite{wen2022transformers}.

In this paper we explore the use of Transformers for on-line signature verification, in which signatures are acquired with pen tablet devices able to capture \textit{X} and \textit{Y} spatial coordinates, pen pressure, and timestamps. We investigate four different Transformer configurations: \emph{i)} a Vanilla Transformer encoder \cite{vaswani2017attention}, \emph{ii)} the THAT Transformer successfully applied to activity recognition \cite{li2021two}, \emph{iii)} a Transformer successfully applied to gait recognition \cite{delgado2022exploring, delgado2023mgaitformer}, and \emph{iv)} a novel configuration based on the Temporal and Channel modules proposed in THAT \cite{li2021two}, but with the Transformer encoder proposed in Vanilla \cite{vaswani2017attention}. To obtain a fair comparison with the literature, we evaluate the proposed configurations according to the experimental protocol proposed in the SVC-onGoing competition \cite{tolosana2022svc}. In particular, we compare the results with our recent Time-Aligned Recurrent Neural Network (TA-RNN) \cite{tolosana2021deepsign}. TA-RNN combines the potential of DTW and RNNs to train more robust systems against forgeries.

% The reminder of the paper is organized as follows. In Section \ref{sec:rel} we provide the literature of on-line signature verification and Transformers applied to behavioral biometrics. In Section \ref{sec:exp} we describe our proposed architectures for on-line signature verification and the SVC-onGoing competition used to evaluate them. Finally, we provide the results of our exploring study in Section \ref{sec:res} and draw conclusions in Section \ref{sec:con}.

\section{Methods}

We explore a Siamese architecture with four different Transformer configurations for on-line signature verification. From the original time signals acquired by the device (\emph{X} and \emph{Y} spatial coordinates and pen pressure), we extract the set of $23$ local features proposed in \cite{fierrez2007hmm}, obtaining additional time signals related to velocity, acceleration, geometric aspects of the signature, etc. These time signals are used as input to our Siamese architecture.

% We train our architectures with Binary Cross Entropy, Adam optimizer, learning rate of 0.00025 with exponential decay ($\gamma$ = 0.95). With a dedicated subset of subjects we compute validation at the end of each epoch. We select the model with the lowest EER on validation to perform the two final evaluation tasks of the SVC-onGoing competition.

\subsection{Transformer Configurations}

\begin{figure}[t]
  \begin{center}
  \includegraphics[width=\linewidth]{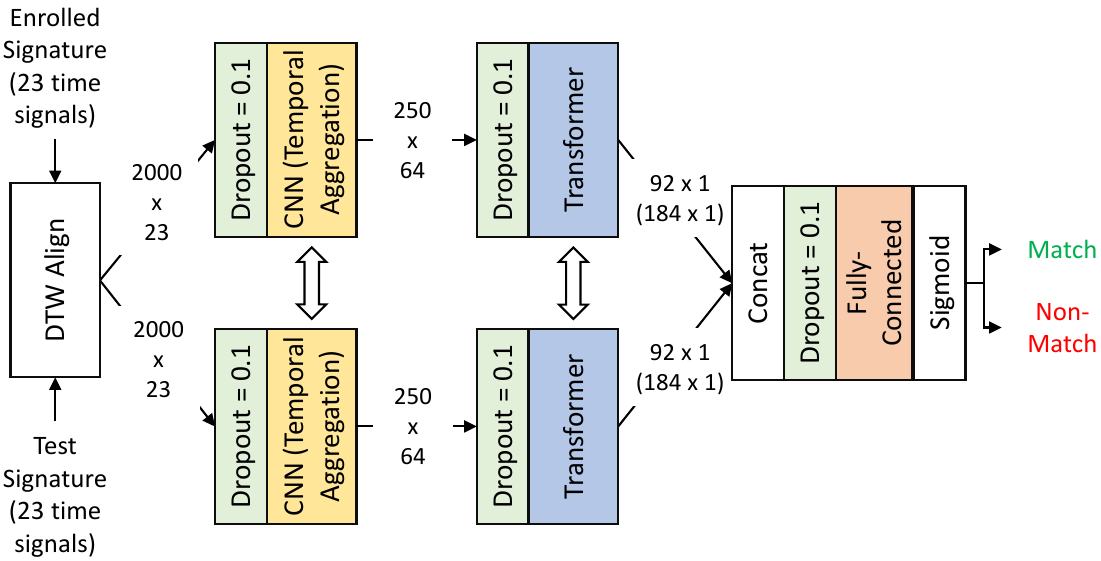}
  \caption{Representation of proposed Siamese architecture based on the combination of CNNs and Transformers.}
  \label{fig:general}
  \end{center}
\end{figure}

Differently from other behavioral biometrics (\emph{e.g.,} gait and keystroke), on-line signatures usually consist in longer sequences. Hence, the processing of time signals with Transformer-based architectures is computationally expensive. Similarly to some approaches proposed for voice analysis \cite{chen2020audio, kong2020sound}, we add convolutional layers prior to Transformers, to aggregate temporal features of the input signals and reduce their dimensionality. A general representation of the proposed architecture is provided in Figure \ref{fig:general}.

The CNN layers consist of a combination of 1D convolutional and MaxPooling layers. Convolutional layers extract temporal features by combining consecutive timesteps and augment the feature size, thanks to the $64$ channels in the output. MaxPooling layers reduce timesteps, making the signals more suitable for Transformers. 

%\begin{table}[]
%\begin{tabular}{c|ccc}

%\textbf{Layer} & \textbf{Kernel size} & \textbf{Output size} & \textbf{Activation} \\ \hline \hline
%1D Conv        & 4                    & $2000 \times 32$     & BN + ReLU           \\ \hline
%1D Conv        & 2                    & $2000 \times 64$     & BN + ReLU           \\ \hline
%1D Max Pool    & 4                    & $500 \times 64$      & -                   \\ \hline
%1D Conv        & 4                    & $500 \times 64$      & BN + ReLU           \\ \hline
%1D Conv        & 2                    & $500 \times 64$      & BN + ReLU           \\ \hline
%1D Max Pool    & 2                    & $250 \times 64$      & -                   \\ 
%\end{tabular}
%\end{table}

Using the Siamese architecture proposed in Figure \ref{fig:general}, we consider four different configurations depending on the Transformer module selected. First, we consider the original Vanilla Transformer encoder \cite{vaswani2017attention}, with Gaussian Range Encoding instead of Positional Encoding, given its suitability with the time series of interest \cite{delgado2022exploring}. The Vanilla encoder processes inputs with size $250 \times 64$ and generates a vector of size $64$ at each timestep. These vectors are processed by a RNN, whose final state of size $92$ is concatenated in the Siamese architecture (see Figure \ref{fig:general}). 

The second and third approaches are based respectively on the THAT Transformer proposed for activity recognition \cite{li2021two} and the Transformer proposed for gait recognition \cite{delgado2022exploring}.

\begin{figure}[t]
\begin{center}
    \includegraphics[width=0.68\linewidth]{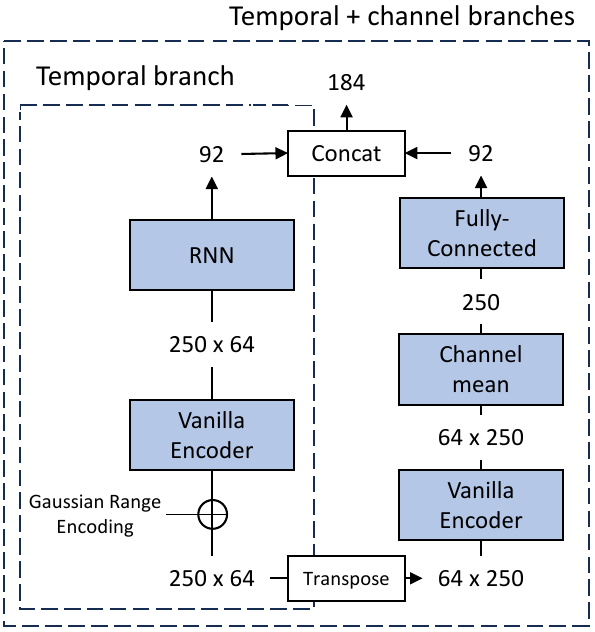}
    
\caption{Representation of the proposed configuration based on the original Vanilla encoder, with Temporal and Channel branches. For Vanilla Encoder (with Temporal branch only) refer to the small box.}
\label{fig:transformer}
\end{center}
\end{figure}

Finally, we explore a novel configuration based on the Temporal and Channel modules proposed in THAT \cite{li2021two}, but with the Transformer encoder proposed in Vanilla \cite{vaswani2017attention}. The Temporal branch is analogous to the one considered in the first configuration. The Channel branch processes inputs with size $64 \times 250$ and generates a vector of size $250$ for each channel. We average these vectors and apply a Fully-Connected layer to reduce the size of the output to $92$ (the same of Temporal branch). The outputs of Temporal and Channel branches are concatenated, being the final vector of size $184$ (Figure \ref{fig:transformer}). 

\section{Experimental Setup}
We evaluate our configurations with the experimental protocol proposed in the SVC-onGoing competition \cite{tolosana2022svc}. Two publicly available datasets are considered in the competition: DeepSignDB \cite{tolosana2021deepsign}, used for development and validation, and SVC2021\_EvalDB \cite{tolosana2022svc}, used for the final evaluation. Different subjects are considered in each dataset. In particular, we focus on the office-like scenario where subjects had to perform signatures using a pen tablet device.

To train our four configurations, we generate random pairs of matching and non-matching signatures from the Development DeepSignDB dataset provided by the SVC-onGoing competition. We consider both random and skilled non-matching pairs for training. Finally, following our previous TA-RNN approach \cite{tolosana2021deepsign}, for each signature comparison (i.e., genuine-genuine or genuine-impostor) we align the $23$ time signals of each signature pair with DTW, zero-padding them to obtain a fixed length of 2,000 time samples for each signature.  

\section{Results}
\label{sec:res}
\setlength{\tabcolsep}{4.2pt}
\begin{table}[]
\caption{EER results obtained in the SVC-onGoing competition for the proposed configurations. R = Random, S = Skilled, O = Overall.}
\begin{small}
\begin{tabular}{c|ccc|ccc}
\multirow{2}{*}{\textbf{Configuration}} & \multicolumn{3}{c}{\textbf{DeepSignDB}} & \multicolumn{3}{c}{\textbf{SVC2021\_EvalDB}} \\
                                & R       & S       & O       & R       & S       & O      \\ \hline \hline
TA-RNN \cite{tolosana2022svc, gonzalez2023introduction}               & 1.87\% & 5.77\%                                                                                                & \textbf{4.31\%}           & 1.91\% & 5.83\%       & 4.08\%           \\ \hline \hline
Vanilla encoder    \cite{vaswani2017attention}     & 3.25\% & 5.92\%                   & 4.64\%     &3.52\% &   4.06\%           & \textbf{3.80\%}           \\ \hline
THAT Transf. \cite{li2021two}          & 4.57\% & 8.02\%                                                        & 6.35\%       & 5.13\% & 6.75\%          & 6.03\%           \\  \hline
Gait Transformer \cite{delgado2022exploring}                   & 3.84\% & 6.42\%                & 5.13\%    & 3.60\% & 4.44\%             & 4.10\%           \\ \hline
\begin{tabular}[c]{@{}c@{}}Vanilla encoder\\ (Temporal + Channel)\end{tabular}         & 4.13\% & 6.60\%         & 5.36\%        & 4.58\% & 4.42\%          & 4.49\%           \\ 
\end{tabular}
\end{small}
\label{tab:results}
\end{table}

The results obtained by evaluating our four Transformer configurations according to the protocol of the SVC-onGoing competition are reported in Table \ref{tab:results}. We consider non-match comparisons made of random signatures (R), skilled signatures (S), and an overall combination of the two (O). Random signatures is the type of impostors that always provide the lowest EER, except in the case of the Vanilla encoder with Temporal and Channel branches, evaluated on SVC2021\_EvalDB that provides 4.42\% EER for skilled non-match comparisons and 4.58\% EER for random ones. 

The Transformer configurations achieve similar performance compared to the TA-RNN previously presented. From the four Transformer configurations explored, we observe that the Vanilla encoder achieves the best overall results, 4.64\% EER and 3.80\% EER for the DeepSignDB and SVC2021\_EvalDB, respectively. Focusing on the SVC2021\_EvalDB dataset considered for the final evaluation of the competition, we can observe how the Vanilla encoder achieves a relative improvement of 7\% in comparison to the TA-RNN approach (4.08\% EER), showing a better generalisation ability to the new scenarios not considered in training.

The results of Table \ref{tab:results} show how more complex configurations do not improve the results obtained in evaluation. Overall EERs raise from 4.64\% to 5.36\% for DeepSignDB and from 3.80\% to 4.49\% for SVC2021\_EvalDB when we add the channel branch to the configuration based on Vanilla encoder. Similar results apply to the other two configurations considered, with the Transformer proposed for gait recognition that only get closer to the best performances, with overall EERs of 5.13\% and 4.10\% in DeepSignDB and SVC2021\_EvalDB evaluations.

%\begin{figure*}[h]
%  \centering
%  \includegraphics[width=\linewidth]{images/metrics.pdf}
%  \caption{1907 Franklin Model D roadster. Photograph by Harris \&
%    Ewing, Inc. [Public domain], via Wikimedia
%    Commons. (\url{https://goo.gl/VLCRBB}).}
%  \Description{A woman and a girl in white dresses sit in an open car.}
%\end{figure*}

%\section{Conclusions}
%\label{sec:con}

%In this exploratory study we applied Transformer architectures to the task of online signature verification. We have investigated different architectures and the obtained results are promising. The length of these temporal signals require to combine Transformers with other components of Deep Learning architectures. In particular, the application of prelimiary CNNs allows to get closer to the old results obtained in the SVC Ongoing competition. Interestingly, the simplest architecture provide s the best results. This gives hints for the future development of the theme. Novel architectures may consider the evaluation of windows (subsets) of the entire signal, with the possibility of combining the output of each window to improve the accuracy of predictions. 

\newpage

\begin{acks}
This project has received funding from the European Union’s Horizon 2020 research and innovation programme under the Marie Skłodowska-Curie grant agreement No 860813 - TReSPAsS-ETN. With support also from projects INTER-ACTION (PID2021-126521OB-I00 MICINN/FEDER) and HumanCAIC (TED2021-131787B-I00 MICINN).
\end{acks}

\bibliographystyle{ACM-Reference-Format}
\bibliography{mybib}

\end{document}